# A New Concept of Modular Parallel Mechanism for Machining Applications

Damien Chablat and Philippe Wenger
Institut de Recherche en Communications et Cybernétique de Nantes[1]
1, rue de la Noë, 44321 Nantes, France
Damien.Chablat@irccyn.ec-nantes.fr

## Abstract

The subject of this paper is the design of a new concept of modular parallel mechanisms for three, four or five-axis machining applications. Most parallel mechanisms are designed for three- or six-axis machining applications. In the last case, the position and the orientation of the tool are coupled and the shape of the workspace is complex. The aim of this paper is to use a simple parallel mechanism with two-degree-of-freedom (dof) for translation motions and to add one or two legs to add one or two-dofs for rotation motions. The kinematics and singular configurations are studied for each mechanism.

## 1   Introduction

Parallel kinematic machines (PKM) are commonly claimed to offer several advantages over their serial counterparts, like high structural rigidity, high dynamic capacities and high accuracy [1]. Thus, PKM are interesting alternative designs for high-speed machining applications.

The first industrial application of PKMs was the Gough platform, designed in 1957 to test tyres [2]. PKMs have then been used for many years in flight simulators and robotic applications [3] because of their low moving mass and high dynamic performances [1]. This is why parallel kinematic machine tools attract the interest of most researchers and companies. Since the first prototype presented in 1994 during the IMTS in Chicago by Gidding&Lewis (the VARIAX), many other prototypes have appeared.

To design a parallel mechanism, two important problems must be solved. The first one is the singular configurations, which can be located inside the workspace. For a six-dof parallel mechanism, like the Gough-Stewart platform, the location of the singular configurations is very difficult to characterize and can change as a function of small variations of the design parameters [3]. The second problem is the non-homogeneity of its performance indices (condition number, stiffness...) throughout the workspace [1]. To the authors' knowledge, only one parallel mechanism is isotropic throughout the workspace [4] but its stiffness is insufficient to be used in machining applications because its legs are subject to bending. Unfortunately, this concept is limited to three-dof mechanisms and cannot be extended to four or five-dof parallel mechanisms.

Numerous papers deal with the design of parallel mechanisms [4,5]. However, there is a lack of four- or five-dof parallel mechanisms, which are especially needed for machining applications [6].

To decrease the cost of industrialization of new PKM and to reduce the problems of design, a modular strategy can be applied. The translation and rotation motions can be divided into two separated parts to produce a mechanism where the direct kinematic problem is decoupled. This simplification yields also some simplifications in the definition of the singular configurations.

The organization of this paper is as follows. Next section presents design problems of parallel mechanisms. The kinematic description and singularity analysis of the parallel mechanism used, are reported in section 3. Section 4 is devoted to design of two new architectures of parallel mechanisms with one or two dofs of rotation.

---

## 2 About parallel kinematic machines

### 2.1 General remarks

In a PKM, the tool is connected to the base through several kinematic chains or legs that are mounted in parallel. The legs are generally made of telescopic struts with fixed foot points (Figure 1a), or fixed length struts with moveable foot points (Figure 1b).

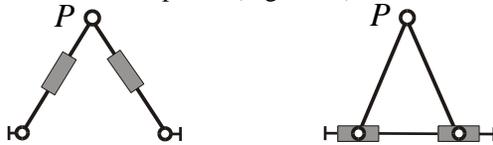

Figure 1a: A bipod PKM       Figure 1b: A biglide PKM

For machining applications, the second architecture is more appropriate because the masses in motion are lower. The linear joints can be actuated by means of linear motors or by conventional rotary motors with ball screws.
A classification of the legs suitable to produce motions for parallel kinematic machines is given by [6] with their degrees of freedom and constraints. The connection of identical or different kinematic legs permits the authors to define two-, three-, four- and five-dof parallel mechanisms. However, it is not possible to remove one leg from a four-dof to produce a three-dof mechanism because no modular approach is used.

### 2.2 Singularities

The singular configurations (also called singularities) of a PKM may appear inside the workspace or at its boundaries. There are two types of singularities [7]. A configuration where a finite tool velocity requires infinite joint rates is called a serial singularity. These configurations are located at the boundary of the workspace. A configuration where the tool cannot resist any effort and in turn, becomes uncontrollable is called a parallel singularity. Parallel singularities are particularly undesirable because they induce the following problems (i) a high increase in forces in joints and links, that may damage the structure, and (ii) a decrease of the mechanism stiffness that can lead to uncontrolled motions of the tool though actuated joints are locked.

Figures 2a and 2b show the singularities for the biglide mechanism of Fig. 1b. In Fig. 2a, we have a serial singularity. The velocity amplification factor along the vertical direction is null and the force amplification factor is infinite.

Figure 2b shows a parallel singularity. The velocity amplification factor is infinite along the vertical direction and the force amplification factor is close to zero. Note that a high velocity amplification factor is not necessarily desirable because the actuator encoder resolution is amplified and thus the accuracy is lower.

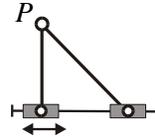    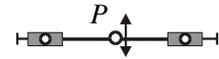

Figure 2a: A serial singularity       Figure 2b: A parallel singularity

The determination of the singular configurations for two-dof mechanisms is very simple; conversely, for a six-dof mechanism like Gough-Stewart platform, a mechanism with six-dof, the problem is very difficult [3]. With a modular architecture, when position and orientation of the mobile platform are decoupled, the determination of singularities is easier.

### 2.3 Kinetostatic performance of parallel mechanism

Various performance indices have been devised to assess the kinetostatic performances of serial and parallel mechanisms. The literature on performance indices is extremely rich to fit in the limits of this paper [9] (service angle, dexterous workspace and manipulability…). The main problem of these performance indices is that they do not take into account the location of the tool frame. However, the Jacobian determinant depends on this location [9] and this location depends on the tool used.
Another problem is that to the authors' knowledge there is no parallel mechanism, suitable for machining, for which the kinetostatic performance indices are constant throughout the workspace (like the condition number or the stiffness…). For a serial three-axis machine tool, a motion of an actuated joint yields the same motion of the tool (the transmission factors are equal to one). For a parallel machine, these motions are generally not equivalent. When the mechanism is close to a parallel singularity, a small joint rate can generate a large velocity of the tool. This means that the positioning accuracy of the tool is lower in some directions for some configurations close to parallel singularities because the encoder resolution is amplified. In addition, a high velocity amplification factor in one direction is equivalent to a loss of stiffness in this direction. The manipulability ellipsoid of the Jacobian matrix of robotic manipulators was defined two decades ago [8]. Unfortunately, this concept is quite difficult to apply when the tool frame can produce rotation and translation motions. Indeed, in this case, the Jacobian matrix is not homogeneous [9].

The first way to solve this problem is its normalization by computing its characteristic length [9]. The second one is to change the form of the Jacobian matrix. The first part of the mechanism for translational motion is optimized using homogeneous matrix. Then, the part dedicated to rotation motion can be optimized using the method introduced by [10].

## 3 Kinematics of mechanisms for translation motions.

Figure 3 shows a PKM with two dofs. The output body is connected to the linear joints through a set of two parallelograms of equal lengths $L = A_i B_i$, so that it can move only in translation.

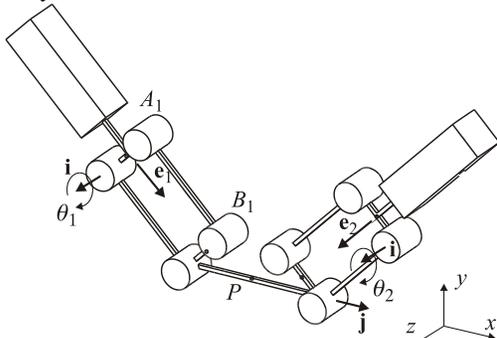

Figure 3: Parallel mechanism with two-dof

The two legs are *PPa* identical chains, where *P* and *Pa* stand for Prismatic and Parallelogram joints, respectively. This mechanism can be optimized to have a workspace whose shape is close to a square workspace and the velocity amplification factors are bounded [11].

The joint variables $\rho_1$ and $\rho_2$ are associated with the two prismatic joints. The output variables are the Cartesian coordinates of the tool center point $P = [x\ y]^T$. To control the orientation of the reference frame attached to *P*, two parallelograms can be used, which also increase the rigidity of the structure, Figure 3.

To produce the third translational motion, it is possible to place orthogonally a third prismatic joint. This one can be located as in the case of Figure 4. Another solution is to use the Orthoglide mechanism, an isotropic three-dof mechanism [12]. The choice between these two solutions depends on the main application of the milling machine. For example, for aeronautical pieces, the solution of fig. 4 is more appropriate because long and fine pieces are built. Conversely, for rapid prototyping of compact parts, we can choose the Orthoglide.

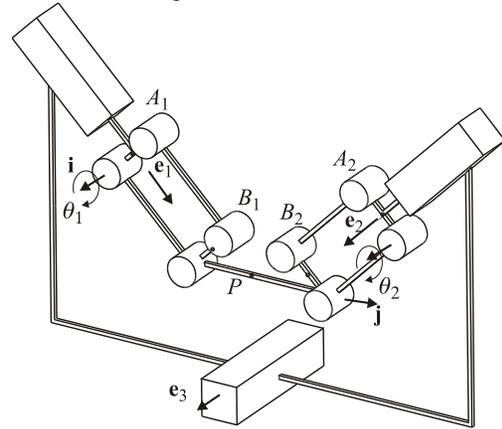

Figure 4: Hybrid mechanism with three-dof

The velocity $\dot{\mathbf{p}}$ of point *P* can be expressed in two different ways. By traversing the closed loop $(A_1 B_1 P - A_2 B_2 P)$ in two possible directions, we obtain

$$\dot{\mathbf{p}} = \dot{\mathbf{a}}_1 + \dot{\theta}_1 \mathbf{i} \times (\mathbf{b}_1 - \mathbf{a}_1) \tag{1a}$$

$$\dot{\mathbf{p}} = \dot{\mathbf{a}}_2 + \dot{\theta}_2 \mathbf{i} \times (\mathbf{b}_2 - \mathbf{a}_2) \tag{1b}$$

where $\mathbf{a}_1$, $\mathbf{b}_1$, $\mathbf{a}_2$ and $\mathbf{b}_2$ represent the position vectors of the points $A_1$, $B_1$, $A_2$ and $B_2$, respectively. Moreover, the velocities $\dot{\mathbf{a}}_1$ and $\dot{\mathbf{a}}_2$ of $A_1$ and $A_2$ are given by $\dot{\mathbf{a}}_1 = \mathbf{e}_1 \dot{\rho}_1$ and $\dot{\mathbf{a}}_2 = \mathbf{e}_2 \dot{\rho}_2$, respectively.

For an isotropic configuration to exist where the velocity amplification factors are equal to one, we must have $\mathbf{e}_1 . \mathbf{e}_2 = 0$ [11] (Figure 5). Two square useful workspaces can be used. The first one has horizontal and vertical sides. The second one has oblique sides but its size is higher.

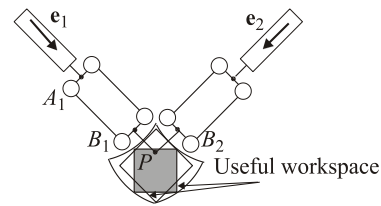

Figure 5: Cartesian workspace and isotropic configuration

We would like to eliminate the two passive joint rates $\dot{\theta}_1$ and $\dot{\theta}_2$ from Eqs. (1a-b), which we do upon dot-multiply the former by $(\mathbf{b}_1 - \mathbf{a}_1)^T$ and the latter by $(\mathbf{b}_2 - \mathbf{a}_2)^T$, thus obtaining

$$(\mathbf{b}_1 - \mathbf{a}_1)^T \dot{\mathbf{p}} = (\mathbf{b}_1 - \mathbf{a}_1)^T \mathbf{e}_1 \dot{\rho}_1 \qquad (2a)$$

$$(\mathbf{b}_2 - \mathbf{a}_2)^T \dot{\mathbf{p}} = (\mathbf{b}_2 - \mathbf{a}_2)^T \mathbf{e}_2 \dot{\rho}_2 \qquad (2b)$$

Equations (2a-b) can be cast in vector form, namely $\mathbf{A}\dot{\mathbf{p}} = \mathbf{B}\dot{\boldsymbol{\rho}}$, with $\mathbf{A}$ and $\mathbf{B}$ denoted, respectively, as the parallel and serial Jacobian matrices,

$$\mathbf{A} \equiv \begin{bmatrix} (\mathbf{b}_1 - \mathbf{a}_1)^T \\ (\mathbf{b}_2 - \mathbf{a}_2)^T \end{bmatrix} \quad \mathbf{B} \equiv \begin{bmatrix} (\mathbf{b}_1 - \mathbf{a}_1)^T \mathbf{e}_1 & 0 \\ 0 & (\mathbf{b}_2 - \mathbf{a}_2)^T \mathbf{e}_2 \end{bmatrix}$$

where $\dot{\boldsymbol{\rho}}$ is defined as the vector of actuated joint rates and $\dot{\mathbf{p}}$ is the velocity of point $P$, i.e.,

$$\dot{\boldsymbol{\rho}} = \begin{bmatrix} \dot{\rho}_1 \\ \dot{\rho}_2 \end{bmatrix} \text{ and } \dot{\mathbf{p}} = \begin{bmatrix} \dot{x} \\ \dot{y} \end{bmatrix}$$

When $\mathbf{A}$ and $\mathbf{B}$ are not singular, we obtain the relations,

$$\dot{\mathbf{p}} = \mathbf{J}\dot{\boldsymbol{\rho}} \text{ with } \mathbf{J} = \mathbf{A}^{-1}\mathbf{B}$$

Parallel singularities occur whenever the lines ($A_1B_1$) and ($A_2B_2$) are colinear, i.e. when $\theta_1 - \theta_2 = k\pi$, for $k = 1,2,...$. Serial singularities occur whenever $\mathbf{e}_1 \perp \mathbf{b}_1 - \mathbf{a}_1$ or $\mathbf{e}_2 \perp \mathbf{b}_2 - \mathbf{a}_2$. In [12], the range limits are defined to avoid these two singularities in using suitable bounds on the velocity factor amplification.

## 4 Kinematics of mechanisms for translation and rotation motions

The aim of this section is to define the kinematics of two mechanisms with one and two dofs of rotation, respectively. To be modular, the direct kinematic problem must be decoupled between position and orientation equations. A decoupled version of the Gough-Stewart Platform exists but it is very difficult to build because three spherical joints must coincide [13]. Thus, it cannot be used to perform milling applications.

The main idea of the proposed architecture is to attach a new body with the tool frame to the mobile platform of the two dofs defined in the previous section. The new joint admits one or two-dof according to the prescribed tasks.

### 4.1 Kinematics of a spatial parallel mechanism with one-dof of rotation

To add one-dof on the mechanism defined in section 3, we introduce one revolute joint between the previous mobile platform and the tool frame. Only one leg is necessary to hold the tool frame in position. Figure 6 shows the mechanism obtained with two translational dofs and one rotational dof.

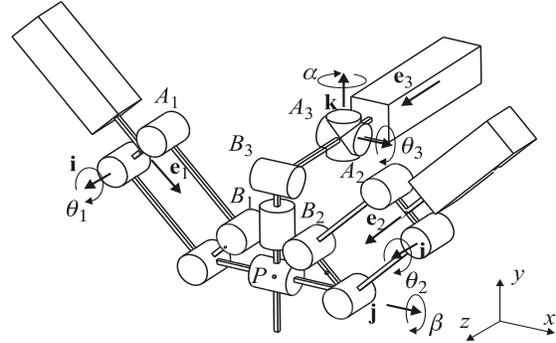

Figure 6: Parallel mechanism with two-dof of translation and one-dof of rotation

The architecture of the leg added is *PUU* where *P* and *U* stand for Prismatic and Universal joints, respectively [6]. The new prismatic joint is located orthogonaly to the first two prismatic joints. This location can be easily justified because on this configuration, i.e. when $\mathbf{b}_1 - \mathbf{a}_1 \perp \mathbf{b}_2 - \mathbf{a}_2$ and $\mathbf{b}_3 - \mathbf{a}_3 \perp \mathbf{b}_3 - \mathbf{p}$, the third leg is far away from serial and parallel singularities.

Let $\dot{\boldsymbol{\rho}}$ be referred to as the vector of actuated joint rates and $\dot{\mathbf{p}}$ as the velocity vector of point $P$,

$$\dot{\boldsymbol{\rho}} = [\dot{\rho}_1 \; \dot{\rho}_2 \; \dot{\rho}_2]^T \text{ and } \dot{\mathbf{p}} = [\dot{x} \; \dot{y}]^T$$

Due to the architecture of the two-dof mechanism and the location of *P*, its velocity on the z-axis is equal to zero. $\dot{\mathbf{p}}$ can be written in three different ways by traversing the three chains $A_iB_iP$,

$$\dot{\mathbf{p}} = \dot{\mathbf{a}}_1 + \dot{\theta}_1 \mathbf{i} \times (\mathbf{b}_1 - \mathbf{a}_1) \qquad (3a)$$

$$\dot{\mathbf{p}} = \dot{\mathbf{a}}_2 + \dot{\theta}_2 \mathbf{i} \times (\mathbf{b}_2 - \mathbf{a}_2) \qquad (3b)$$

$$\dot{\mathbf{p}} = \dot{\mathbf{a}}_3 + (\dot{\theta}_3 \mathbf{j} + \dot{\alpha}\mathbf{k}) \times (\mathbf{b}_3 - \mathbf{a}_3) + \dot{\beta}\mathbf{j} \times (\mathbf{p} - \mathbf{b}_3) \qquad (3c)$$

where $\mathbf{a}_i$ and $\mathbf{b}_i$ represent the position vectors of the points $A_i$ and $B_i$ for $i = 1,2,3$, respectively. Moreover, the velocities $\dot{\mathbf{a}}_1$, $\dot{\mathbf{a}}_2$ and $\dot{\mathbf{a}}_3$ of $A_1$, $A_2$ and $A_3$ are given by $\dot{\mathbf{a}}_1 = \mathbf{e}_1\dot{\rho}_1$, $\dot{\mathbf{a}}_2 = \mathbf{e}_2\dot{\rho}_2$ and $\dot{\mathbf{a}}_3 = \mathbf{e}_3\dot{\rho}_3$, respectively.

We want to eliminate the passive joint rates $\dot{\theta}_i$ and $\dot{\alpha}$ from Eqs. (3a-c), which we do upon dot-multiplying Eqs. (3a-c) by $\mathbf{b}_i - \mathbf{a}_i$,

$$(\mathbf{b}_1 - \mathbf{a}_1)^T \dot{\mathbf{p}} = (\mathbf{b}_1 - \mathbf{a}_1)^T \mathbf{e}_1 \dot{\rho}_1 \qquad (4a)$$

$$(\mathbf{b}_2 - \mathbf{a}_2)^T \dot{\mathbf{p}} = (\mathbf{b}_2 - \mathbf{a}_2)^T \mathbf{e}_2 \dot{\rho}_2 \qquad (4b)$$

$$(\mathbf{b}_3 - \mathbf{a}_3)^T \dot{\mathbf{p}} = (\mathbf{b}_3 - \mathbf{a}_3)^T \mathbf{e}_3 \dot{\rho}_3 \\ + (\mathbf{b}_3 - \mathbf{a}_3)^T \dot{\beta}\mathbf{j} \times (\mathbf{p} - \mathbf{b}_3) \qquad (4c)$$

Equations (4a-c) can be cast in vector form, namely,

$\mathbf{t} = \mathbf{J}\dot{\boldsymbol{\rho}}$ with $\mathbf{J} = \mathbf{A}^{-1}\mathbf{B}$ and $\mathbf{t} = [\dot{x}\ \dot{y}\ \dot{\beta}]^T$

where $\mathbf{A}$ and $\mathbf{B}$ are the parallel and serial Jacobian matrices, respectively,

$$\mathbf{A} \equiv \begin{bmatrix} (\mathbf{b}_1 - \mathbf{a}_1)^T & 0 \\ (\mathbf{b}_2 - \mathbf{a}_2)^T & 0 \\ (\mathbf{b}_3 - \mathbf{a}_3)^T & -(\mathbf{b}_3 - \mathbf{a}_3)^T \mathbf{j} \times (\mathbf{p} - \mathbf{b}_3) \end{bmatrix}$$

$$\mathbf{B} \equiv \begin{bmatrix} (\mathbf{b}_1 - \mathbf{a}_1)^T \mathbf{e}_1 & 0 & 0 \\ 0 & (\mathbf{b}_2 - \mathbf{a}_2)^T \mathbf{e}_2 & 0 \\ 0 & 0 & (\mathbf{b}_3 - \mathbf{a}_3)^T \mathbf{e}_3 \end{bmatrix}$$

There are two new singularities when one leg is added. The first one is a parallel singularity when $(\mathbf{b}_3 - \mathbf{a}_3)^T \mathbf{j} \times (\mathbf{p} - \mathbf{b}_3) = 0$, i.e., when the lines $(A_3B_3)$ and $(B_3P)$ are colinear, and the second one is a serial singularity when $(\mathbf{b}_3 - \mathbf{a}_3)^T \mathbf{e}_3 = 0$, i.e., $\mathbf{a}_3 - \mathbf{b}_3 \perp \mathbf{e}_3$. However, these singular configurations are simple and can be avoided by proper limits on the actuated joints.

### 4.2　Kinematics of a spatial mechanism with two-dof of rotation

To add two dofs on the mechanism defined in section 3, we introduce a universal joint, between the previous mobile platform and the tool frame. Two legs are necessary to hold the tool frame in position and we use the same archtitecture than for the previous mechanism, i.e. the *PUU* mechanism. Figure 7 depicts a parallel mechanism with two translational dofs and two rotational dofs.

Let $\dot{\boldsymbol{\rho}}$ be referred to as the vector of actuated joint rates and $\dot{\mathbf{p}}$ as the velocity vector of point P,

$\dot{\boldsymbol{\rho}} = [\dot{\rho}_1\ \dot{\rho}_2\ \dot{\rho}_3\ \dot{\rho}_4]^T$ and $\dot{\mathbf{p}} = [\dot{x}\ \dot{y}\ ]^T$

$\dot{\mathbf{p}}$ can be written in four different ways by traversing the three chains $A_iB_iP$,

$\dot{\mathbf{p}} = \dot{\mathbf{a}}_1 + \dot{\theta}_1\mathbf{i} \times (\mathbf{b}_1 - \mathbf{a}_1)$ (5a)

$\dot{\mathbf{p}} = \dot{\mathbf{a}}_2 + \dot{\theta}_2\mathbf{i} \times (\mathbf{b}_2 - \mathbf{a}_2)$ (5b)

$\dot{\mathbf{p}} = \dot{\mathbf{a}}_3 + (\dot{\theta}_3\mathbf{i}_3 + \dot{\alpha}_3\mathbf{k}) \times (\mathbf{b}_3 - \mathbf{a}_3) + (\dot{\beta}\mathbf{j} + \dot{\gamma}\mathbf{i}) \times (\mathbf{p} - \mathbf{b}_3)$ (5c)

$\dot{\mathbf{p}} = \dot{\mathbf{a}}_4 + (\dot{\theta}_4\mathbf{i}_4 + \dot{\alpha}_4\mathbf{k}) \times (\mathbf{b}_4 - \mathbf{a}_4) + (\dot{\beta}\mathbf{j} + \dot{\gamma}\mathbf{i}) \times (\mathbf{p} - \mathbf{b}_4)$ (5d)

where $\mathbf{a}_i$ and $\mathbf{b}_i$ represent the position vectors of the points $A_i$ and $B_i$, respectively, for $i = 1,2,3,4$. Moreover, the velocities $\dot{\mathbf{a}}_1$, $\dot{\mathbf{a}}_2$, $\dot{\mathbf{a}}_3$ and $\dot{\mathbf{a}}_4$ of $A_1$, $A_2$, $A_3$ and $A_4$ are given by $\dot{\mathbf{a}}_1 = \mathbf{e}_1\dot{\rho}_1$, $\dot{\mathbf{a}}_2 = \mathbf{e}_2\dot{\rho}_2$, $\dot{\mathbf{a}}_3 = \mathbf{e}_3\dot{\rho}_3$ and $\dot{\mathbf{a}}_4 = \mathbf{e}_4\dot{\rho}_4$, respectively.

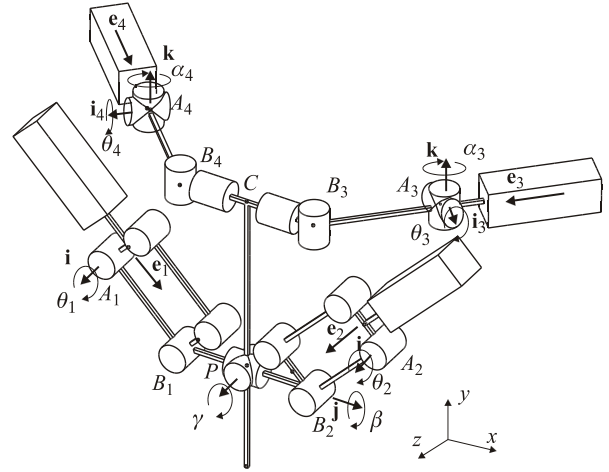

Figure 7: Parallel mechanism with two-dof of translation and two-dof of rotation

We want to eliminate the passive joint rates $\dot{\theta}_i$ and $\dot{\alpha}_i$ from Eqs. (6a-d), which we do upon dot-multiplying Eqs. (5a-d) by $\mathbf{b}_i - \mathbf{a}_i$,

$(\mathbf{b}_1 - \mathbf{a}_1)^T \dot{\mathbf{p}} = (\mathbf{b}_1 - \mathbf{a}_1)^T \mathbf{e}_1 \dot{\rho}_1$ (7a)

$(\mathbf{b}_2 - \mathbf{a}_2)^T \dot{\mathbf{p}} = (\mathbf{b}_2 - \mathbf{a}_2)^T \mathbf{e}_2 \dot{\rho}_2$ (7b)

$(\mathbf{b}_3 - \mathbf{a}_3)^T \dot{\mathbf{p}} = (\mathbf{b}_3 - \mathbf{a}_3)^T \mathbf{e}_3 \dot{\rho}_3$
$\qquad + (\mathbf{b}_3 - \mathbf{a}_3)^T (\dot{\beta}\mathbf{j} + \dot{\gamma}\mathbf{i}) \times (\mathbf{p} - \mathbf{b}_3)$ (7c)

$(\mathbf{b}_4 - \mathbf{a}_4)^T \dot{\mathbf{p}} = (\mathbf{b}_4 - \mathbf{a}_4)^T \mathbf{e}_4 \dot{\rho}_4$
$\qquad + (\mathbf{b}_4 - \mathbf{a}_4)^T (\dot{\beta}\mathbf{j} + \dot{\gamma}\mathbf{i}) \times (\mathbf{p} - \mathbf{b}_4)$ (7d)

Equations (8a-d) can be cast in vector form, namely,

$\mathbf{t} = \mathbf{J}\dot{\boldsymbol{\rho}}$ with $\mathbf{J} = \mathbf{A}^{-1}\mathbf{B}$ and $\mathbf{t} = \begin{bmatrix} \dot{x} & \dot{y} & \dot{\beta} & \dot{\gamma} \end{bmatrix}^T$

where $\mathbf{A}$ and $\mathbf{B}$ are the parallel and serial Jacobian matrices, respectively,

$$\mathbf{A} \equiv \begin{bmatrix} (\mathbf{b}_1 - \mathbf{a}_1)^T & 0 & 0 \\ (\mathbf{b}_2 - \mathbf{a}_2)^T & 0 & 0 \\ (\mathbf{b}_3 - \mathbf{a}_3)^T & -(\mathbf{b}_3 - \mathbf{a}_3)^T \mathbf{j} \times (\mathbf{p} - \mathbf{b}_3) & -(\mathbf{b}_3 - \mathbf{a}_3)^T \mathbf{i} \times (\mathbf{p} - \mathbf{b}_3) \\ (\mathbf{b}_4 - \mathbf{a}_4)^T & -(\mathbf{b}_4 - \mathbf{a}_4)^T \mathbf{j} \times (\mathbf{p} - \mathbf{b}_4) & -(\mathbf{b}_4 - \mathbf{a}_4)^T \mathbf{i} \times (\mathbf{p} - \mathbf{b}_4) \end{bmatrix}$$

and

$$\mathbf{B} \equiv \begin{bmatrix} (\mathbf{b}_1 - \mathbf{a}_1)^T \mathbf{e}_1 & 0 & 0 & 0 \\ 0 & (\mathbf{b}_2 - \mathbf{a}_2)^T \mathbf{e}_2 & 0 & 0 \\ 0 & 0 & (\mathbf{b}_3 - \mathbf{a}_3)^T \mathbf{e}_3 & 0 \\ 0 & 0 & 0 & (\mathbf{b}_4 - \mathbf{a}_4)^T \mathbf{e}_4 \end{bmatrix}$$

When two legs are added, the same singular configurations as the previous mechanism occur, a parallel singularity occurs when the lines $(A_3B_3)$, $(A_4B_4)$ and $(CP)$ are coplanar, and a serial singularity occurs when $(\mathbf{b}_3 - \mathbf{a}_3)^T \mathbf{e}_3 = 0$, *i.e.*, $\mathbf{a}_3 - \mathbf{b}_3 \perp \mathbf{e}_3$ or $(\mathbf{b}_4 - \mathbf{a}_4)^T \mathbf{e}_4 = 0$, *i.e.*, $\mathbf{a}_4 - \mathbf{b}_4 \perp \mathbf{e}_4$.

### 4.3  Discussion

The problem of the optimal design of the two mechanisms defined in the previous sections is not addressed in this paper. In future works, for both mechanisms, it is possible to define a configuration for which the Jacobian matrix is isotropic. This result permits us to define the location of the prismatic joints and to define the condition length to normalize the Jacobian matrix. A method to define the range limits is explained in [10], *via* a distance to the isotropic configuration.

## 5  Conclusions

In this paper, a new class of modular mechanisms is introduced with two, three, four and five dofs. All the actuated joints are prismatic joints, which can be actuated by means of linear motors or by conventional rotary motors with ball screws. The topology of the legs used to add one or two dof is the same. Only three types of joints are used, *i.e.*, prismatic, revolute and universal joints. All the singularities are characterized easily because position and orientation are decoupled for the direct kinematic problem and can be avoid by proper design. In the future, for the modular architecture, the lengths of the legs as well as their positions will be optimized, to take into account the velocity amplification factors.

## 6  Acknowledgments

This research was partially supported by the CNRS (Project ROBEA "Machine à Architecture compleXe"). The authors would like to thank Mr. Caro for his valuable remarks on this paper.